\documentclass[11pt]{article}
\PassOptionsToPackage{table}{xcolor}

\usepackage[preprint]{acl}

\usepackage{times}
\usepackage{booktabs,tabularx,array}
\usepackage{latexsym}
\usepackage{hyperref} %
\usepackage[capitalize,noabbrev]{cleveref}

\usepackage[T1]{fontenc}

\usepackage[utf8]{inputenc}

\usepackage{microtype}

\usepackage{inconsolata}

\usepackage{graphicx}
\usepackage{booktabs}
\usepackage[table]{xcolor}

\usepackage{cuted}
\usepackage{capt-of}
\usepackage{enumitem}

\title{VELD-QA: A Large-Scale Dataset for Visual Evidence Grounding and Question Answering in Long Documents}

\author{\normalfont
 Zhivar Sourati\textsuperscript{1,2},
 Marianne Menglin Liu\textsuperscript{2},
 Rongguang Wang\textsuperscript{2},
 Zheng Wang\textsuperscript{2},
 Yazhe Hu\textsuperscript{2},
\\
 Lan Guo\textsuperscript{2},
 Yuying Wang\textsuperscript{2},
 Mengqing Guo\textsuperscript{2},
 Bhavesh Dnyaneshwar Gore\textsuperscript{3},
 Shubham Pathak\textsuperscript{3},
\\
 Ramapriya Ranganath\textsuperscript{3},
 Sandeep Sharma\textsuperscript{3},
 Tao Sheng\textsuperscript{2},
 Sujith Ravi\textsuperscript{2},
 Dan Roth\textsuperscript{2,4}
\\
 \textsuperscript{1}University of Southern California,
 \textsuperscript{2}Oracle AI,
 \textsuperscript{3}Turing Enterprise Inc,
 \textsuperscript{4}University of Pennsylvania
}

\begin{document}
\maketitle
\begin{abstract}

Existing long-document QA benchmarks rarely combine large-scale human-authored questions with detailed visual evidence annotations, limiting diagnosis of evidence localization and interpretation failures. We present \textbf{VELD-QA}, a large-scale benchmark for \textbf{V}isual \textbf{E}vidence grounding and question answering in \textbf{L}ong \textbf{D}ocuments. It contains 15,000 fully human-authored question-answer pairs across 5,859 real-world PDFs totaling 372,445 pages, balancing chart, table, and complex-layout tasks. Human annotations identify supporting pages and evidence elements and describe cited visual content, enabling separate retrieval and interpretation evaluation. We evaluate nine vision-language models under four evidence settings and seven retrieval-augmented and agentic systems. Our analysis identifies bottlenecks in evidence localization and interpretation. Retrieval-based methods substantially underperform the oracle setting with ground-truth supporting pages; full-document access does not close the gap. Localization alone is insufficient: oracle accuracy peaks at 81.7\%, highlighting persistent visual interpretation and reasoning challenges. Enriching inputs with our visual-element annotations narrows the performance gap by 17--24\%, demonstrating value beyond evaluation and motivating data enrichment for multimodal document QA. Our code is available at \href{https://github.com/oracle-samples/veld-qa}{https://github.com/oracle-samples/veld-qa}.

\end{abstract}

\section{Introduction}

Real-world documents convey information through text, tables, charts, and figures, with page layout providing structural cues that help readers locate content and interpret its relationships. Answering questions about these documents requires finding relevant evidence and understanding it in context, which becomes increasingly challenging in long documents where evidence is sparse or dispersed across pages. Multimodal document question answering (DocQA) therefore depends on both effective evidence retrieval and accurate interpretation. Reflecting these demands, evaluation has expanded from individual pages and charts to long documents, alongside growing attention to locating and extracting supporting evidence \cite{mathew2021docvqa, masry2022chartqa, ma2024mmlongbenchdoc, deng2025longdocurl, dong2025mmdocir, feng2026docscope}. Evaluating retrieval and interpretation together, while distinguishing their limitations, is essential for identifying bottlenecks and building reliable document understanding systems.

\begin{figure*}[!t]
  \centering
  \setlength{\abovecaptionskip}{3pt}
  \setlength{\belowcaptionskip}{0pt}
  \includegraphics[
    width=\textwidth,
    trim={8pt 4pt 8pt 4pt},
    clip
  ]{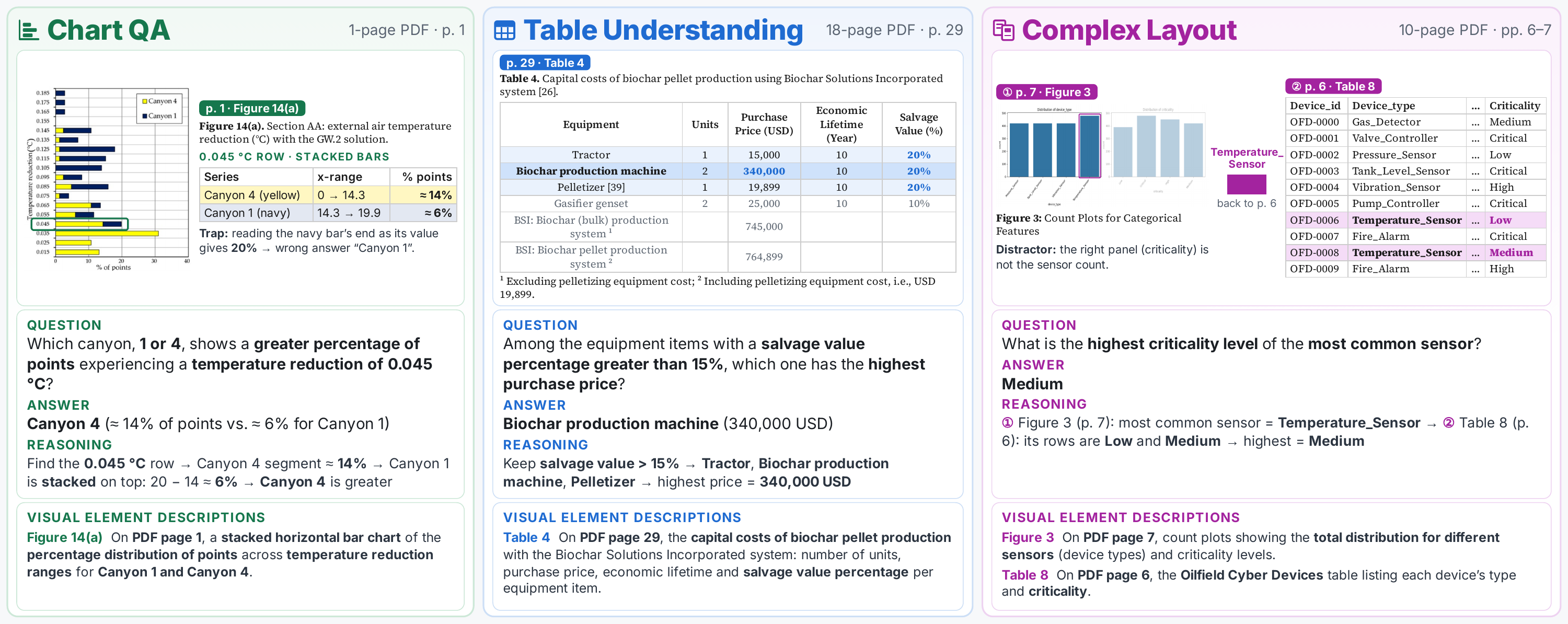}
\caption{
  \textbf{VELD-QA} comprises 15,000 human-authored QA examples across three
  task categories: \textbf{Chart QA}, \textbf{Table Understanding}, and
  \textbf{Complex Layout}. Each example shows, from top to bottom: page-tagged evidence excerpts with
answer-relevant content highlighted; the question, answer, and reasoning trace;
and descriptions of the cited visual elements.
}
  \label{fig:veld-qa-examples}
\end{figure*}

Nevertheless, current benchmarks overlook key challenges in reliable document evaluation. Large-scale datasets must cover diverse layouts and sources, from academic papers to business reports, with chart, table, and complex QA tasks spanning descriptive, comparative, and analytical questions of varying difficulty. Contamination threatens fair evaluation; human collection from diverse sources and independent annotation can reduce the risk of complete examples appearing in training data. Missing links between questions, answers, and visual evidence also compromise accuracy and obscure retrieval versus interpretation errors. Our results in \Cref{subsec:qtypes} and prior work \cite{opsahl2026officeqa} show that ground-truth supporting pages or full documents alone do not ensure accurate answers. Visual-element annotations help narrow this gap (\Cref{subsec:annotations}), highlighting data enrichment as a promising direction for multimodal document QA.

We introduce VELD-QA, a multimodal DocQA benchmark with 15,000 human-authored questions over 5,859 real-world PDFs totaling 372,445 pages, with 5,000 questions each for chart, complex-layout, and table understanding. Chart questions use single-page excerpts; the 10,000 complex-layout and table questions cover 3,617 PDFs averaging 102 pages and reaching 1,798 pages. Across diverse document genres, answers range from factual responses to comparative and analytical explanations.

Human annotators create all examples and reference answers, with supporting pages and logical evidence elements, including charts, tables, figures, and text, annotated for nearly all examples. These annotations distinguish page retrieval from within-page evidence identification and support evaluations targeting both. Document-disjoint training, development, and test splits support controlled evaluation. \Cref{sec:dataset-construction} details construction, annotation, and dataset statistics.

We establish baselines for nine vision-language models on 1,452 test questions under four evidence settings: question only, gold pages, top-3 visual retrieval \citep{faysse2025colpali}, and all pages. We also compare seven retrieval-augmented and agentic systems \citep{sourati2026ladrag,han2025mdocagent,jain2025simpledoc,sun2025docagent,zhu2026doclens,borchmann2026madqa}.
Even with the gold pages, the best model reaches only 81.7\% accuracy, and tables, especially from financial reports, remain the hardest. The remaining failures are of two kinds. First, models do not reliably locate the evidence: retrieving three pages costs every model 2--19 points relative to gold pages, reading all pages helps only the strongest models, and agentic search recovers only part of this gap at a much higher cost. Second, models do not fully read the evidence they are given: adding the annotated table worksheets to the same gold pages raises table accuracy by 13--14 points.

Our contributions are threefolds:
\begin{itemize}[leftmargin=*,topsep=2pt,itemsep=1pt,parsep=0pt]
    \item A large benchmark of real-world PDFs from diverse domains, featuring heterogeneous coverage of charts, tables, and complex layouts, low question repetition, and a broad range of tasks spanning single-page interpretation and long-document evidence retrieval.
    \item Human-annotated evidence and visual descriptions, with document-disjoint splits, enable separate evaluation of retrieval, enrichment, and end-to-end QA. Enriching inputs with these descriptions improves QA accuracy through richer evidence, demonstrating value beyond evaluation and the promise of data enrichment for multimodal document QA.
    \item We evaluate nine vision-language models under four evidence settings and seven retrieval-augmented and agentic systems, identifying evidence localization and interpretation as the two key bottlenecks in long-document QA.
\end{itemize}

\begin{table*}[!t]
\centering
\scriptsize
\setlength{\tabcolsep}{4pt}
\renewcommand{\arraystretch}{1.10}

\newcommand{\veldYes}{\mbox{\usefont{U}{pzd}{m}{n}\char51}}
\newcommand{\veldNo}{\mbox{\usefont{U}{pzd}{m}{n}\char55}}

\caption{
Comparison of \textbf{VELD-QA} with long-document QA benchmarks.
Human QA indicates fully human-authored questions.
Element descriptions are annotations linked to cited visual elements,
and cross-element measures questions requiring evidence from multiple element types.
}
\label{tab:veld-comparison}

\vspace{-2pt}

\resizebox{\textwidth}{!}{%
\begin{tabular}{@{}lcrrrrcccc@{}}
\toprule
\textbf{Benchmark}
& \textbf{Release}
& \multicolumn{2}{c}{\textbf{Dataset Scale}}
& \multicolumn{2}{c}{\textbf{Document Scale}}
& \multicolumn{2}{c}{\textbf{Answer Evidence}}
& \multicolumn{2}{c}{\textbf{Annotation}}
\\

\cmidrule(lr){3-4}
\cmidrule(lr){5-6}
\cmidrule(lr){7-8}
\cmidrule(l){9-10}

&
&
\textbf{\#Docs}
&
\textbf{\#QA}
&
\shortstack{\textbf{Avg.}\\\textbf{Pages}}
&
\shortstack{\textbf{Avg.}\\\textbf{Tokens}}
&
\shortstack{\textbf{Cross-page}\\\textbf{(\%)}}
&
\shortstack{\textbf{Cross-element}\\\textbf{(\%)}}
&
\shortstack{\textbf{Fully Human}\\\textbf{QA}}
&
\shortstack{\textbf{Human Visual}\\\textbf{Element Descriptions}}
\\

\midrule

SlideVQA~\citep{tanaka2023slidevqa}
& 2023-01
& 2,619
& 14,484
& 20
& 2,030.5
& ---
& ---
& \veldYes
& \veldNo
\\

MMLongBench-Doc~\citep{ma2024mmlongbenchdoc}
& 2024-07
& 135
& 1,082
& 47.5
& 21,214.1
& 33.7
& 22.6
& \veldYes
& \veldNo
\\

SPIQA~\citep{pramanick2024spiqa}
& 2024-07
& 118
& 666
& 10
& 5,000
& ---
& ---
& \veldNo
& \veldNo
\\

M-LongDoc~\citep{chia2025mlongdoc}
& 2024-11
& 180
& 851
& 210.8
& 120,988.0
& 0
& ---
& \veldNo
& \veldNo
\\

LongDocURL~\citep{deng2025longdocurl}
& 2024-12
& 396
& 2,325
& 85.6
& 43,622.6
& 52.9
& 37.1
& \veldNo
& \veldNo
\\

MMDocIR~\citep{dong2025mmdocir}
& 2025-01
& 313
& 1,658
& 65.1
& 35,800
& 18.88
& ---
& \veldYes
& \veldNo
\\

SciMDR~\citep{chen2026scimdr}
& 2026-03
& 200
& 907
& ---
& ---
& ---
& ---
& \veldYes
& \veldNo
\\

OfficeQA~\citep{opsahl2026officeqa}
& 2026-03
& 697
& 133
& 127.7
& 70,200
& ---
& ---
& \veldYes
& \veldNo
\\

DocScope~\citep{feng2026docscope}
& 2026-05
& 273
& 1,124
& 51.3
& 24,561
& 57.7
& ---
& \veldNo
& \veldNo
\\

VinQA~\citep{jang2026vinqa}
& 2026-06
& ---
& 1,605
& ---
& ---
& 64.17
& ---
& \veldNo
& \veldNo
\\

FinMRAGBench~\citep{yang2026finmragbench}
& 2026-07
& 723
& 887
& 133.5
& 73,400
& 88.73
& ---
& \veldNo
& \veldNo
\\

XL-DocBench~\citep{wei2026xldocbench}
& 2026-07
& 331
& 1,519
& 297.1
& 227,463.2
& 72.6
& ---
& \veldNo
& \veldNo
\\

\midrule

\textbf{VELD-QA (Ours)}
& \textbf{2026-10}
& \textbf{5,859}
& \textbf{15,000}
& \textbf{63.6}
& \textbf{44,700}
& \textbf{10.25}
& \textbf{18.92}
& \veldYes
& \veldYes
\\

\bottomrule
\end{tabular}%
}

\vspace{-4pt}
\end{table*}

\section{Related Work}
\label{sec:related-work}

\subsection{Visual and Long-Document QA Benchmarks}
Document question-answering benchmarks span different visual content and input scopes. DocVQA~\citep{mathew2021docvqa} evaluates document-image reading, ChartQA~\citep{masry2022chartqa} emphasizes visual and numerical chart reasoning, and SlideVQA~\citep{tanaka2023slidevqa} introduces evidence selection across slide images. MMLongBench-Doc~\citep{ma2024mmlongbenchdoc}, LongDocURL~\citep{deng2025longdocurl}, and M-LongDoc~\citep{chia2025mlongdoc} extend evaluation to long PDFs with heterogeneous visual evidence, numerical reasoning, localization, and open-ended answering. XL-DocBench~\citep{wei2026xldocbench} further studies extra-long contexts, while FinMRAGBench~\citep{yang2026finmragbench} evaluates evidence gathering across financial reports. Scientific resources such as SPIQA~\citep{pramanick2024spiqa} and SciMDR~\citep{chen2026scimdr} also provide large training corpora. OfficeQA Pro~\citep{opsahl2026officeqa} evaluates end-to-end grounded reasoning over U.S. Treasury Bulletins, including retrieval, numerical analysis, and the effects of document representation. Unlike benchmarks focused primarily on end-to-end QA accuracy, VELD-QA enables fine-grained diagnosis of evidence retrieval and visual interpretation. Its human-verified evidence descriptions and table transcriptions further support controlled evaluation of how visual enrichment improves downstream QA.

\subsection{Evidence Grounding and Evaluation}
Answer correctness and evidence support capture different aspects of document understanding. Supporting-page annotations enable retrieval evaluation, while MMDocIR~\citep{dong2025mmdocir} additionally evaluates retrieval of layout regions. DocScope~\citep{feng2026docscope} decomposes reasoning into page selection, region grounding, fact extraction, and answering. VinQA~\citep{jang2026vinqa} studies explanatory responses interleaved with cited visual elements, and MMLongCite~\citep{zhou2025mmlongcite} evaluates citation-grounded generation over long multimodal contexts. These efforts motivate separate assessment of evidence localization, answer quality, and attribution. Our annotations identify supporting pages and visual elements and describe the cited visual content, enabling separate evaluation of evidence localization and interpretation. Element-level descriptions are important because locating the correct page alone does not ensure accurate interpretation of its visual evidence. Enriching model inputs with these descriptions improves QA accuracy.

\section{Dataset Construction and Curation}
\label{sec:dataset-construction}

\subsection{Corpus and Source Selection}
\begin{figure*}[!t]
  \centering
  \setlength{\abovecaptionskip}{3pt}
  \setlength{\belowcaptionskip}{0pt}
    \includegraphics[width=\textwidth]{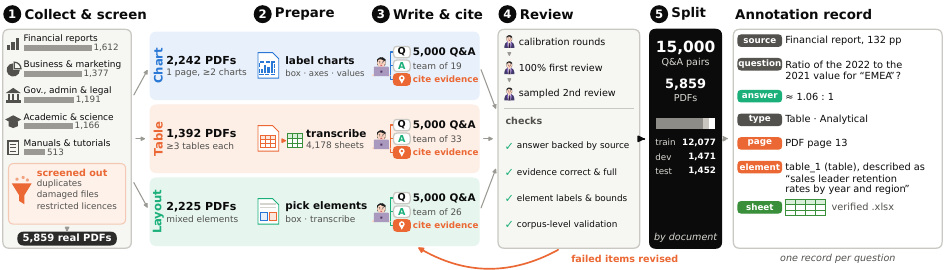}
\caption{
  \textbf{VELD-QA construction.} Real PDFs from five genres are collected and screened (1); each task then follows its own lane, where annotators prepare the content that questions will cite (2) and write each question and answer with its supporting evidence (3). All items pass a shared review with calibration, complete first-level and sampled second-level review, and quality checks, with failed items returned for revision (4), and the data are split by document (5). Right: an example annotation record from the test set. Every question, answer, and evidence citation is written by a human.
}
  \label{fig:veld-qa-pipeline}
\end{figure*}

We construct a multimodal document QA benchmark comprising 15,000 question--answer pairs across three equally sized subsets: \emph{Chart QA}, \emph{Table QA}, and \emph{Complex-layout QA}. We design these subsets to evaluate graphical interpretation, tabular reasoning, and evidence integration within heterogeneous documents. Each instance includes a question, a reference answer, a source PDF, and supporting evidence annotations. \Cref{fig:veld-qa-pipeline} outlines the construction process, and \Cref{tab:dataset-statistics} summarizes the benchmark.

We source documents through targeted web searches and screen candidates for relevance, duplication, technical integrity, accessibility, and reuse restrictions. Document sourcing and screening involve 2 contributors for Chart QA, 5 for Table QA, and 2 for Complex-layout QA. We manually verify eligibility against task-specific requirements: multiple charts on a single page for Chart QA, clearly delineated tables in multi-page documents for Table QA, and heterogeneous pages combining textual and visual elements for Complex-layout QA. Our sources span research publications, financial and business reports, government documents, and manuals. We provide detailed selection criteria in Appendix~\ref{app:construction-details}.

\begin{table}[t]
\caption{Statistics for VELD-QA benchmark.}
\label{tab:dataset-statistics}
\centering
\small
\begin{tabular}{@{}lr@{}}
\toprule
\textbf{Statistic} & \textbf{Number} \\
\midrule
\multicolumn{2}{@{}l}{\textbf{Document}} \\
\quad PDF files & 5,859 \\
\quad Distinct PDF byte contents & 5,585 \\
\quad Total PDF pages & 372,445 \\
\quad Avg. pages per PDF file & 63.6 \\
\quad Spreadsheet workbooks & 1,392 \\
\midrule
\multicolumn{2}{@{}l}{\textbf{Question \& Answer}} \\
\quad Total QA pairs & 15,000 \\
\quad Avg. question words & 16.7 \\
\quad Max. question words & 69 \\
\midrule
\multicolumn{2}{@{}l}{\textbf{Task Type} (all QA pairs)} \\
\quad Chart understanding & 5,000 (33.3\%) \\
\quad Table understanding & 5,000 (33.3\%) \\
\quad Complex layout & 5,000 (33.3\%) \\
\midrule
\multicolumn{2}{@{}l}{\textbf{Question Type} (chart and table; $n=10{,}000$)} \\
\quad Descriptive & 2,816 (28.2\%) \\
\quad Comparative & 3,251 (32.5\%) \\
\quad Analytical & 3,933 (39.3\%) \\
\midrule
\multicolumn{2}{@{}l}{\textbf{Evidence Type} (complex layout; $n=5{,}000$)} \\
\quad Text & 2,145 (42.9\%) \\
\quad Table & 1,832 (36.6\%) \\
\quad Chart & 1,582 (31.6\%) \\
\quad Figure & 414 (8.3\%) \\
\midrule
\multicolumn{2}{@{}l}{\textbf{Number of Evidence Pages} (table and layout; $n=10{,}000$)} \\
\quad Single-page evidence & 8,462 (84.6\%) \\
\quad Multi-page evidence & 1,538 (15.4\%) \\
\midrule
\multicolumn{2}{@{}l}{\textbf{Number of Evidence Types} (complex layout; $n=5{,}000$)} \\
\quad Single type & 4,054 (81.1\%) \\
\quad Multiple types & 946 (18.9\%) \\
\quad Avg. cited elements, multiple types & 2.12 \\
\bottomrule
\end{tabular}

\end{table}

\subsection{Human Annotation and Evidence Grounding}
We prepare task-specific content representations and source-grounded questions with annotation teams of 19 for Chart QA, 33 for Table QA, and 26 for Complex-layout QA. For Chart QA, we annotate chart types, complexity, axes, legends, headers, and data values to support questions about individual charts and relationships across charts. For Table QA, we manually transcribe three selected tables per source document into separate worksheets and verify their numerical and structural fidelity against the PDF. The consolidated subset contains 1,392 workbooks with 4,178 worksheets. For Complex-layout QA, a parser locates candidate pages, and annotators then select the evidence and record element types, page references, citation counts, ordinal positions, and descriptions.

We organize question annotation around descriptive retrieval, comparison, and analytical reasoning, including table filtering and aggregation and questions spanning multiple content elements or pages. We associate each answer with citations to its supporting content. Comparative and analytical questions account for 71.84\% of the 10,000 chart and table instances; complex-layout records retain evidence-type rather than question-type labels. After normalizing page references, we identify 808 table questions (16.16\%) and 730 complex-layout questions (14.60\%) with multi-page citations, and 946 complex-layout questions (18.92\%) with multiple evidence types. We report these as measures of annotated evidence coverage, not proof that every cited element is necessary. The evidence is also sparse: 40\% of the multi-page PDFs exceed 100 pages, and the gold pages make up a median of only 1.7\% of a document's pages.

\subsection{Review and Quality Control}
We establish a shared understanding of the annotation guidelines through calibration exercises and apply complete first-level human review, sampled secondary review, and iterative correction. We use automated checks for screening, completeness, and consistency, supplemented by automated review of chart questions, answers, citations, and coverage. Human reviewers make the final decisions on visual suitability, evidence attribution, and answer correctness. We return failed or contested annotations for revision and address recurring errors through guideline updates and targeted feedback. Finally, we audit the consolidated corpus for structural validity and confirm that all questions and answers are nonempty and all referenced PDFs and workbooks are present. We discuss review procedures in Appendix~\ref{app:construction-details}.

\subsection{Question Diversity and Task Coverage}
\label{sec:question-diversity}

LLM-mediated content can converge on recurring linguistic and reasoning patterns, potentially limiting the diversity needed for flexible problem-solving~\citep{sourati2026homogenizing}. In benchmarks, such concentration can leave capabilities underexplored and steer benchmark-driven development toward a narrow set of skills, even when individual questions and answers are correct. We therefore examine whether VELD-QA's scale is accompanied by diversity in both question content and task demands.

\begin{figure}[t]
\centering
\includegraphics[width=\columnwidth]{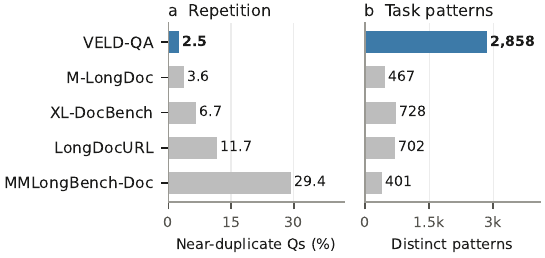}
\caption{Question diversity across document-QA benchmarks. (a) Estimated percentage of questions with a semantically similar question among others from the same benchmark (MPNet embeddings, cosine similarity $\geq$ 0.8). (b) Distinct combinations of reasoning operations, constraints, and requested answer forms in each benchmark's annotated questions; these counts grow with the number of questions (size-matched comparison in the text).}
\label{fig:question-diversity}
\end{figure}

\paragraph{Low repetition at scale.} With comparison size held constant, VELD-QA has an estimated semantic repetition rate of 2.55\%, compared with 29.43\% for MMLongBench-Doc and 11.69\% for LongDocURL, with relative reductions of 91.3\% and 78.2\%, respectively (\Cref{fig:question-diversity}a). Its rate is also lower than those of M-LongDoc and XL-DocBench, indicating that its 15,000 questions offer substantial scale without concentrating heavily on closely similar questions.

\paragraph{A broad inventory of task patterns.} Using GPT-5.4 to annotate question text, we define a task pattern as a combination of reasoning operations, constraints, and requested answer form. VELD-QA contains 2,858 distinct patterns, the largest inventory among the compared benchmarks (\Cref{fig:question-diversity}b). Because pattern counts grow with the number of questions, we also compare at a matched size of 733 annotated questions: VELD-QA covers 15.6\% and 10.1\% more patterns than MMLongBench-Doc and LongDocURL, but fewer than M-LongDoc and XL-DocBench. These findings highlight the value of investing in human-authored questions at scale: VELD-QA combines low repetition with broad task-pattern coverage, helping limit concentration on recurring demands while providing a varied foundation for evaluating and guiding model development.

\section{Experimental Setup}
\label{sec:experimental-setup}

\paragraph{Evaluation scope.}
We split the benchmark at the PDF level into training, development, and test sets. The development and test sets contain 1,471 and 1,452 questions, respectively. All reported results are on the test set.

\paragraph{Models.}
We evaluate nine frontier and open-source vision-language models: Gemma 3 12B IT \citep{team2025gemma}, MUSE Glimmer 30B \citep{meta2026museglimmer}, Gemini 2.5 Flash and Pro \citep{comanici2025gemini}, Grok 4.20 \citep{xai2026grok420}, GPT-5.5 \citep{openai2026gpt55}, GPT-6 Astra \citep{openai2026gpt6astra}, and Claude Sonnet 5.5 \citep{anthropic2026sonnet55} and Opus 5.5 \citep{anthropic2026opus55}.

\paragraph{Evidence-access settings.}
We compare four settings that vary the document evidence available to the model. \emph{Question only} provides the question without document content and serves as a sanity check for performance without visual evidence to rule out possible model memorization of the document knowledge. \emph{Gold pages} provides images of the annotated supporting pages, isolating answer generation from page retrieval. \emph{Top-3 RAG} retrieves the top three pages within the source PDF and supplies their images to the answering model. We use ColQwen2 for Gemma, MUSE, Gemini, Grok, and GPT-5.5, and ColPali \citep{faysse2025colpali} for GPT-6 Astra and the Claude models. \emph{All pages} processes the entire PDF without relevance-based page selection. Documents with up to eight pages are processed directly; longer documents are processed in page batches, followed by synthesis of the intermediate findings. For single-page chart documents, the visual settings provide access to the same page. Reference answers are used only for evaluation.

\subsection{Agentic Document Understanding}
\label{sec:agentic-document-understanding}
In addition to top-3 RAG, we compare six retrieval-augmented and agentic systems using a shared GPT-5.4 backbone~\citep{openai2026gpt54}: MDocAgent~\citep{han2025mdocagent}, SimpleDoc~\citep{jain2025simpledoc}, DocAgent~\citep{sun2025docagent}, the BM25-MLLM agent of MADQA~\citep{borchmann2026madqa}, DocLens~\citep{zhu2026doclens}, and LAD-RAG~\citep{sourati2026ladrag}. These systems span fixed retrieval, multi-agent verification, and adaptive evidence acquisition. We retain method-specific prompts and budgets, with configurations and implementation adaptations detailed in Appendix~\ref{app:agentic-system-configurations}. Gold-page and question-only runs with the same backbone and answer prompt provide reference settings. We report per-question cost using token usage across all calls.

\subsection{Evaluation Metric}
\label{sec:evaluation-metric}
We report \emph{accuracy}: whether an answer agrees with the reference answer in meaning. Because answers range from single values to multi-sentence explanations, exact string matching is too strict, so we use Gemini 2.5 Flash as a LLM judge with a fixed rubric, without revealing the answering model or its evidence-access setting. The judge labels each answer as correct, partially correct, or incorrect; partially correct answers count half, and accuracy is the resulting mean in percent. We report document-level bootstrap 95\% confidence intervals and paired differences on identical questions. An independent GPT-5.5 assessment of 3,323 stratified answers yields strong agreement (quadratic-weighted $\kappa=0.91$) and similar rankings (Kendall $\tau=0.91$). We detail the rubric, uncertainty estimation, and judge validation in Appendix~\ref{app:evaluation-protocol}.

\section{Benchmarking Results}
\label{sec:results}

\subsection{Oracle evidence improves accuracy but leaves substantial gaps}
\label{subsec:main}

\Cref{tab:main-results} reports all nine models under the four evidence settings. Without the document, no model exceeds 21.4\% accuracy, and table accuracy is only 1.6--4.2\%, indicating that memorization or prior knowledge alone is insufficient for strong performance on VELD-QA. Supplying the annotated supporting pages raises accuracy to between 39.1\% (Gemma 3 12B IT) and 81.7\% (Claude Opus 5.5). Even with this perfect evidence, a large gap remains: 9.2\% of test questions are not fully answered by any model, and table questions are the hardest task for every model (at most 76.0\%, compared with 88.9\% on complex-layout questions). Model rankings are stable across evidence settings (Kendall $\tau$ = 0.83--0.89). Open-weight models are competitive when given the right evidence: MUSE Glimmer 30B matches Gemini 2.5 Pro with gold pages (73.9 vs.\ 74.0\%) but falls to 53.2\% when it must read every page. 

\begin{table}[t]
\centering
\caption{Accuracy (\%) on the VELD-QA test set under four evidence settings. Top-3 RAG retrieves pages with ColQwen2 (\textsuperscript{Q}) or ColPali (\textsuperscript{P}); the other settings use no retriever.}
\label{tab:main-results}
\small
\setlength{\tabcolsep}{3pt}
\resizebox{\columnwidth}{!}{\begin{tabular}{@{}lrrrr|rr@{}}
\toprule
Model & \shortstack{Question\\only} & \shortstack{Gold\\pages} & \shortstack{Top-3\\RAG} & \shortstack{All\\pages} & \shortstack{Gold\\$-$RAG} & \shortstack{All\\$-$RAG} \\
\midrule
\rowcolor{gray!15}
\multicolumn{7}{@{}l}{\textit{Open-weight}} \\
Gemma 3 12B IT & 15.1 & 39.1 & 37.1\textsuperscript{Q} & 28.2 & +2.1 & $-$8.8 \\
MUSE Glimmer 30B & 13.2 & 73.9 & 66.4\textsuperscript{Q} & 53.2 & +7.5 & $-$13.3 \\
\rowcolor{gray!15}
\multicolumn{7}{@{}l}{\textit{Proprietary}} \\
Gemini 2.5 Flash & 11.6 & 70.5 & 51.7\textsuperscript{Q} & 46.1 & +18.7 & $-$5.6 \\
Gemini 2.5 Pro & 14.2 & 74.0 & 59.4\textsuperscript{Q} & 56.0 & +14.6 & $-$3.4 \\
Grok 4.20 & 16.5 & 64.6 & 60.8\textsuperscript{Q} & 56.6 & +3.9 & $-$4.1 \\
GPT-5.5 & 20.0 & 76.0 & 71.3\textsuperscript{Q} & 72.0 & +4.7 & +0.7 \\
GPT-6 Astra & 19.9 & 80.2 & 72.5\textsuperscript{P} & 75.4 & +7.5 & +2.8 \\
Claude Sonnet 5.5 & 18.1 & 81.2 & 74.9\textsuperscript{P} & 78.9 & +6.3 & +4.0 \\
Claude Opus 5.5 & \textbf{21.4} & \textbf{81.7} & \textbf{74.9}\textsuperscript{P} & \textbf{79.1} & +6.9 & +4.3 \\
\bottomrule
\end{tabular}}

\end{table}

\begin{table*}[t]
\centering
\caption{Retrieval-augmented and agentic systems with a shared GPT-5.4 backbone. We report accuracy (\%) and paired differences from top-3 RAG in percentage points, with 95\% confidence intervals. Top-3 RAG and gold-page input use identical answer prompts and page rendering.}
\label{tab:agentic}
\small
\setlength{\tabcolsep}{5pt}
\resizebox{0.8\textwidth}{!}{%
\begin{tabular}{@{}lrrrrr@{}}
\toprule
System  & Accuracy & $\Delta$ vs.\ RAG [95\% CI] & $\Delta$ Chart & $\Delta$ Table & $\Delta$ Layout \\
\midrule
Gold pages (oracle) & 78.8 & +4.5 [+2.7, +6.2] & $-$1.6 & +10.2 & +5.2 \\
All pages & 72.3 & $-$2.0 [$-$4.2, +0.1] & $-$6.6 & +3.8 & $-$2.9 \\
Top-3 RAG (ColPali) & 74.4 & -- & -- & -- & -- \\
\midrule
LAD-RAG~\citep{sourati2026ladrag} & 75.7 & +1.3 [$-$0.6, +3.1] & $-$1.0 & +4.8 & +0.3 \\
MDocAgent~\citep{han2025mdocagent} & 76.6 & +2.2 [+0.0, +4.2] & $-$1.3 & +8.4 & $-$0.1 \\
SimpleDoc~\citep{jain2025simpledoc} & 77.9 & +3.5 [+1.4, +5.7] & +1.9 & +8.0 & +1.0 \\
DocLens$^{\dagger}$~\citep{zhu2026doclens} & 77.8 & +2.1 [$-$0.1, +4.3] & $-$1.5 & +6.5 & +1.8 \\
DocAgent~\citep{sun2025docagent} & 69.9 & $-$4.5 [$-$7.1, $-$2.3] & $-$8.6 & +0.7 & $-$5.2 \\
MADQA BM25-MLLM~\citep{borchmann2026madqa} & 61.4 & $-$12.9 [$-$16.0, $-$10.3] & $-$26.5 & $-$0.7 & $-$10.4 \\
\bottomrule
\end{tabular}%
}

\end{table*}

\subsection{Retrieval weakens as documents grow and evidence spans pages}
\label{subsec:retrieval}

Because every question is annotated with its supporting pages, VELD-QA can be used to evaluate the retrieval system separately from question answering (\Cref{tab:retrieval}). On non-chart questions, visual retrievers place 83--88\% of the gold pages in their top three, against 63\% for a text retriever. Retrieval degrades exactly where long-document understanding is hard: Recall@3 drops by about 15 points when the evidence spans two or more pages (ColPali: 70.0 vs.\ 85.2\%) and falls steadily with document length (88.2\% for documents of at most 20 pages vs.\ 78.9\% for 101--300 pages).

\begin{table}[t]
\centering
\caption{Page retrieval against the human-annotated evidence pages on non-chart test questions (chart documents are single pages). Recall@$k$: share of gold pages in the top $k$. 
}
\label{tab:retrieval}
\small
\setlength{\tabcolsep}{3pt}
\resizebox{\columnwidth}{!}{\begin{tabular}{@{}lrrrr|rr|rrr@{}}
\toprule
 & \multicolumn{4}{c|}{Recall@$k$} & \multicolumn{2}{c|}{Gold pages} & \multicolumn{3}{c}{Document pages} \\
Retriever & 1 & 3 & 5 & 10 & 1 & $\geq$2 & $\leq$20 & 21--100 & 101--300 \\
\midrule
ColQwen2 & 68.6 & 88.3 & 93.0 & 96.0 & 90.9 & 73.3 & 92.4 & 87.8 & 85.4 \\
ColPali & 63.1 & 82.9 & 90.7 & 94.8 & 85.2 & 70.0 & 88.2 & 82.9 & 78.9 \\
E5 (text) & 40.3 & 63.3 & 72.9 & 83.0 & 66.2 & 46.8 & 76.5 & 63.5 & 53.0 \\
\bottomrule
\end{tabular}}

\end{table}

\subsection{Complete retrieval coverage does not guarantee reliable evidence use}
\label{subsec:distraction}

We find that every model performs worse with retrieved pages than with gold pages (\Cref{tab:main-results}, Gold$-$RAG), but the source of this gap varies across models. For the Claude models and GPT-6 Astra, the gap is small when retrieval includes all gold pages and substantially larger when evidence is missing. In contrast, the Gemini models retain large performance drop even when all gold pages are available, frequently citing the correct page while extracting an incorrect value. These results distinguish two limitations of retrieval-augmented document QA: retrieving the necessary evidence and using it reliably in the presence of additional context. We present the retrieval-coverage analysis and pipeline controls in Appendix~\ref{app:retrieval-distraction}.

\subsection{Agentic gains are task-dependent and computationally costly}
\label{subsec:agentic}

We compare system designs using a shared backbone and the questions answered by every system (\Cref{tab:agentic}). Top-3 RAG achieves 74.4\%, compared with 72.3\% for full-document input and 78.8\% for gold pages. SimpleDoc and MDocAgent improve on RAG by 3.5 and 2.2 points, respectively, whereas the gains from DocLens and LAD-RAG are not statistically reliable. The improvements are concentrated in table understanding task, with no corresponding advantage on chartQA or complex-layout questions; DocAgent and the text-first MADQA agent instead underperform RAG. Thus, additional retrieval and reasoning steps yield selective rather than general improvements, while increasing cost. Gold-page input remains more accurate than every system and has the lowest cost. We report the cost comparison and a detailed analysis of LAD-RAG's evidence selection in Appendix~\ref{app:agentic-gains-cost}.

\section{Beyond Evaluation: Insights from Benchmark Annotations}
\label{sec:analysis}

\begin{figure}[t]
\centering
\includegraphics[width=\columnwidth]{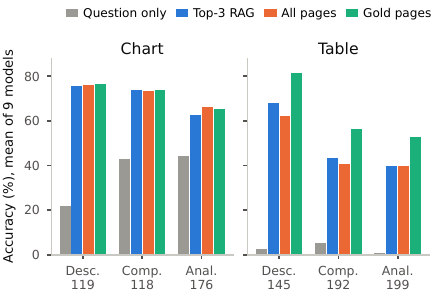}
\caption{Accuracy by question type and evidence setting, averaged over the nine models, for chart and table questions (number of test questions under each label). Comparative and analytical questions are much harder than descriptive ones only on tables.}
\label{fig:qtypes}
\end{figure}

\subsection{Gold evidence does not resolve the table-reasoning bottleneck}
\label{subsec:qtypes}

Comparative and analytical table questions remain substantially harder than descriptive ones even with gold pages (\Cref{fig:qtypes}). Averaged over nine models with top-3 retrieval, they score 24.6 and 28.2 percentage points lower; on charts, only analytical questions are harder (13.0 points). Similar gaps with gold and all pages indicate difficulty interpreting and reasoning over tables that retrieval alone cannot resolve. Tables comprise 37\% of test questions but 62\% of those no model fully answers with gold pages. Without documents, the ordering reverses, partly because binary-choice chart questions permit guessing.

\subsection{Evidence visual annotations improve document understanding}
\label{subsec:annotations}
\begin{figure}[t]
\centering
\includegraphics[width=\columnwidth]{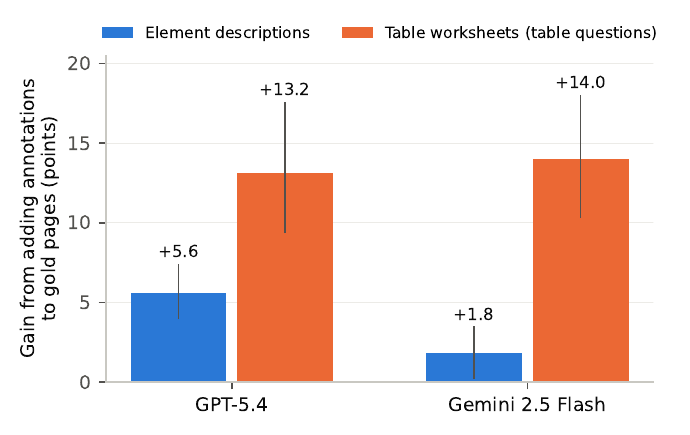}
\caption{Gain in accuracy from adding VELD-QA's annotations of the cited elements, as text, to the same gold pages (paired). Error bars: 95\% CIs. The annotations are question-specific gold evidence: a diagnostic of reading, not a deployable system.}
\label{fig:annotations}
\end{figure}

VELD-QA annotates supporting visual elements with natural-language descriptions and extracted table worksheets. These provide an oracle probe of reading: fully reading gold pages should leave little benefit from added text. Descriptions and worksheets improve GPT-5.4's table accuracy by 5.6 and 13.2 percentage points; worksheets improve Gemini 2.5 Flash by 14.0 points (\Cref{fig:annotations}; 95\% CIs shown). Gains peak on comparative tables (+17.4) and persist with retrieved pages (+22.7 for worksheets). Descriptions alone yield only 16.6\% and largely contain answers for just 2.4\% of questions, suggesting limited shortcut potential. These annotations quantify and provide supervision for reading gaps, especially in tables, and serve as reference enrichments for evaluating document-processing methods.

\subsection{Document genre annotations shape retrieval and reading difficulty}
\label{subsec:genre}

\begin{figure}[t]
\centering
\includegraphics[width=\columnwidth]{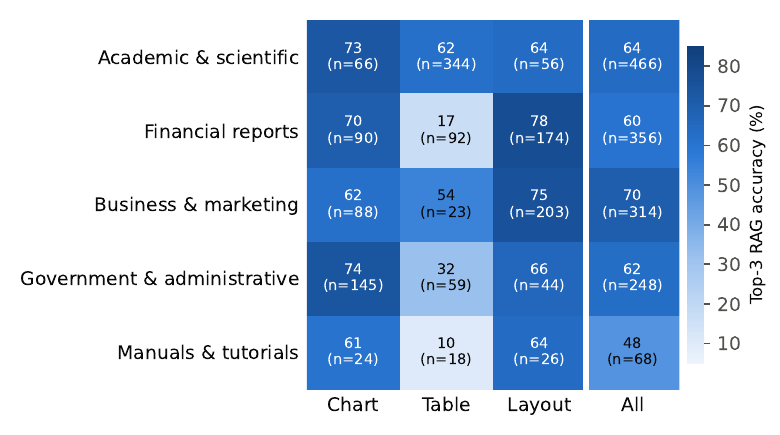}
\caption{Top-3 RAG accuracy (\%) by document genre and task, averaged over the nine models, with the number of test questions per cell. The 22 source categories are grouped into five genres.}
\label{fig:genre}
\end{figure}

We group the 22 source categories into five genres. \Cref{fig:genre} shows top-3 RAG accuracy by genre and task, averaged over the nine models. Genre matters little for charts (61--74\%) but strongly for tables and complex layouts. Tables from financial reports are the hardest part of the benchmark: models reach 17.1\% with retrieved pages and only 28.9\% even with the gold pages, compared with 61.7\% and 74.3\% on tables from academic and scientific documents. The answer is on the page but not read: adding the extracted worksheets to the gold pages raises GPT-5.4 on these questions from 27.7\% to 68.5\%, compared with a 5.7-point gain on academic tables (\Cref{subsec:annotations}). Complex layouts show where retrieval matters: in financial and business reports retrieval costs about ten points (77.8 and 75.4\% vs.\ 87.2\% with gold pages), whereas in academic documents it costs almost nothing (63.7 vs.\ 64.0\%) and reading the layout itself is the bottleneck. Manuals and tutorials are the hardest genre overall (48.4\%, 68 questions).

\section{Conclusion}  

We introduce VELD-QA, a benchmark of 15,000 human-authored questions over diverse real-world PDFs, with balanced coverage of charts, tables, and complex layouts. Its tasks span single-page interpretation and evidence retrieval in long documents. Human-annotated supporting pages, visual-element descriptions, and extracted tables enable separate analysis of retrieval and interpretation, while document-disjoint splits support controlled evaluation. Enriching model inputs with these annotations improves QA accuracy, demonstrating their value beyond evaluation. Our experiments with nine vision-language models under four evidence settings and seven retrieval-augmented and agentic systems reveal persistent challenges in locating and interpreting evidence, particularly in tables. VELD-QA provides a foundation for diagnosing these limitations and developing more reliable, evidence-grounded document QA systems.

\section*{Limitations}

We cover nine vision-language models and seven retrieval-augmented and agentic systems, run with their reported configurations rather than matched compute. Compute-matched comparisons, newer and reasoning-oriented models, document-specialized models, and system variants such as stronger graph builders for LAD-RAG or dedicated table-structure recovery would further map the design space; our results point to multi-page retrieval and table reading as the most promising directions, and our annotation probe suggests that systems which produce element descriptions or table transcriptions themselves could close part of the reading gap. VELD-QA covers English, born-digital PDFs across five genres, and extending it to other languages, scanned and handwritten documents, and formats such as slides and web pages would broaden the evaluation. We report question-only results to show how much can be answered without the document; filtering guessable questions, which mostly come from charts, could yield a harder subset. Finally, each question concerns a single document and is answered in one turn; questions spanning document collections, multi-turn interactions in which users refine their information needs, and answers grounded in localized regions rather than logical elements are natural extensions for which VELD-QA's page- and element-level annotations provide a starting point.

\section*{Ethical Considerations}
VELD-QA supports research on multimodal document understanding, but performance on this benchmark should not be interpreted as evidence of readiness for high-stakes deployment. Our results show that models can miss relevant evidence or misinterpret information even when the correct pages are provided. Such errors, including incorrect numerical extraction and unsupported answers, could have serious consequences in medical, legal, or financial applications. Systems used in these settings require additional validation and appropriate human oversight. Human annotations may also contain errors or subjective judgments, and differences in document genre and task coverage can limit the generalizability of aggregate results. We screen source documents for accessibility, reuse restrictions, and sensitive personal information, but public availability alone does not establish unrestricted reuse. Users should respect applicable permissions and privacy requirements when accessing or redistributing source material. We encourage use of the evidence annotations to audit model behavior rather than treating generated answers or citations as inherently trustworthy.

\bibliography{refs}

\clearpage
\appendix

\section{Additional Dataset Construction Details}
\label{app:construction-details}

\subsection{Source Eligibility and Screening}
We identify candidate PDFs through targeted searches over the designated document categories. We check public accessibility, URL viability, automated-access restrictions, file integrity, page counts, text extractability, and relevance. Our exclusion criteria cover access-controlled or corrupted files, presentation slides, non-substantive or predominantly image-based material, embedded audiovisual content, and sensitive personal information. We manually verify category assignments against document content.

We screen reuse permissions by examining document text, host-site terms, license statements, hyperlinks, and third-party attributions. Our sourcing criteria accept public-domain or approved permissively licensed material and require rejection or further review of sources with restrictive conditions. We separately examine attributed visual assets when their permissions differ from those of the surrounding document. This screening is part of source collection; we do not claim a subsequent independent rights audit of every consolidated file.

We screen for duplicates using URL and title comparisons, normalized document identifiers, identifiers shared across source collections, and chart-title checks where available. We manually review candidate duplicates and source additional documents to address category gaps. We distinguish this screening process from exact-file uniqueness in the consolidated corpus, which we assess separately below.

\subsection{Task-Specific Preparation}
\paragraph{Chart QA.}
We target clear, single-page PDFs containing at least two visible charts and manually assess chart count, clarity, category, and relevance. We annotate chart type and complexity alongside axes, legends, headers, sources, and data points. We construct descriptive, comparative, and analytical questions and attach citations and descriptions linking answers to the selected charts. All PDFs retained in this subset are single-page documents.

\paragraph{Table QA.}
We specify a sourcing target of at least 11 pages and three suitable tables per document, prioritizing numeric tables while allowing text tables in manuals and guidebooks. We manually create spreadsheets and transcribe three distinct tables into separate worksheets. We check values, precision, headers, completeness, gridlines, wrapping, spacing, superscripts, subscripts, and annotation marks. We use structural checks to verify workbook, worksheet, and annotation consistency, but rely on manual comparison with the PDF for content validation. We do not use automated extraction, formatting transfer, or formula-based answer validation as annotation acceptance criteria.

We report the observed corpus separately from the sourcing targets. The retained PDFs span 5--1,138 pages, including six documents shorter than 11 pages. Of the 1,392 workbooks, 1,391 contain three worksheets and one contains five, yielding 4,178 worksheets. We count these retained representations rather than all tables appearing in the source PDFs.

\paragraph{Complex-layout QA.}
We target documents with at least seven pages, including at least two pages that contain four of five element types: titles, text, charts, tables, and figures. Where possible, we seek three qualifying pages to provide additional choices for cross-page questions. We use a parser to suggest candidate pages and manually select meaningful content. We localize evidence through element classification and citations, recording the document, page, element type, citation count, ordinal position, and description. Our annotation interface enforces required fields, while reviewers verify citation counts and ordinals. We piloted automated ordinal checks but did not retain them as a primary control because of limited efficacy. The consolidated PDFs span 7--1,798 pages.

\subsection{Annotator Calibration and Review}
Annotations were created by trained, paid professional annotators organized into task-specific teams of 19, 33, and 26 annotators for Chart, Table, and Complex-layout QA. We introduce annotators to the guidelines, annotation tools, and review rubric through orientation, calibration exercises, and feedback. We conduct two calibration iterations for chart annotation and two calibration batches for table annotation. For complex-layout annotation, we additionally use preliminary annotation tasks, guided practice, and ongoing review, with repeated errors addressed through targeted feedback.

We apply first-level manual review to all chart and table tasks and every complex-layout question--answer pair, with sampled secondary review and periodic spot checks. We also pass chart batches through an automated reviewer that checks chart count, question types, answer accuracy, grammar, citations, descriptions, domain, and coverage. For tables, we manually assess extraction fidelity, question clarity, type alignment, and calculations. For complex layouts, we emphasize distinctions among content types, repeated elements, nested content, citation order, and page numbering.

We resolve failures and disagreements through feedback analysis, correction, and resubmission. Recurring issues inform rubric clarification, additional calibration, and targeted feedback. Because acceptance scores and review samples concern a larger collection, we do not interpret them as independently measured accuracy estimates for the 15,000-question benchmark.

\subsection{Question Types and Evidence Conventions}
\begin{table}[t]
\centering
\small
\begin{tabular}{lrr}
\toprule
Question type & Chart QA & Table QA \\
\midrule
Descriptive & 1,446 & 1,370 \\
Comparative & 1,530 & 1,721 \\
Analytical & 2,024 & 1,909 \\
\bottomrule
\end{tabular}
\caption{Retained question-type annotations. Complex-layout records do not retain this taxonomy. Labels characterize question types rather than calibrated difficulty.}
\label{tab:construction-question-types}
\end{table}

We identify supporting evidence through element or table references, descriptions, and page labels. To measure page coverage, we distinguish printed folios from physical PDF page indices, normalize physical references, expand explicit ranges, and count repeated references once per question. We apply this normalization only when computing statistics and leave the source annotations unchanged. We identify 823 table questions with two table-citation entries and 946 complex-layout questions with multiple evidence types, comprising 919 two-type and 27 three-type instances.

During annotation, we monitored multimodal coverage using questions involving chart or figure interpretation and cross-page coverage using questions with two unique page references. For benchmark analysis, we adopt more explicit definitions: \emph{multiple evidence types} requires at least two annotated types, and \emph{multiple physical pages} requires at least two distinct PDF pages. We distinguish these measures from the indicators used during annotation and do not infer evidence necessity from citation multiplicity alone.

\section{Agentic System Configurations}
\label{app:agentic-evaluation}
\label{app:agentic-system-configurations}
We use GPT-5.4~\citep{openai2026gpt54} as the common answering backbone to compare system designs while holding the backbone fixed. We retain each system's original prompts and method-specific budgets, subject to the settings and adaptations described below. Because evidence and call budgets differ across methods, this comparison does not impose equal computational budgets.

\paragraph{MDocAgent.}
MDocAgent~\citep{han2025mdocagent} combines image, text, and general-purpose agents over three retrieved pages. Its agents examine complementary evidence representations, check intermediate answers through reflection, and synthesize a final response. We use low reasoning effort.

\paragraph{SimpleDoc.}
SimpleDoc~\citep{jain2025simpledoc} uses an iterative retrieval-and-reading controller. It retrieves ten candidate pages, reranks them using page summaries, and can revise the query or inspect additional pages. We allow at most three retrieval rounds, six distinct read pages, and seven query-level model calls.

\paragraph{DocAgent.}
DocAgent~\citep{sun2025docagent} navigates a document outline using reading tools and a reviewer agent. We allow at most 23 model calls per question.

\paragraph{MADQA BM25-MLLM.}
The BM25-MLLM agent of MADQA~\citep{borchmann2026madqa} searches page text with BM25 and reads the returned pages. We allow at most ten calls per question.

\paragraph{DocLens.}
DocLens~\citep{zhu2026doclens} navigates the whole document in chunks using eight sampled navigators, samples eight candidate answers, and adjudicates them into a single response. Following the original method, we sample navigators and candidate answers at temperature 0.7.

\paragraph{LAD-RAG.}
LAD-RAG~\citep{sourati2026ladrag} represents each document as a layout-aware graph of page elements with contextual summaries and section, reference, and community relations. Its answering agent combines semantic search, symbolic graph queries, and visual page search to determine which evidence to acquire. We allow at most eight rounds and six pages. We construct the graphs with Qwen2.5-VL-7B at half image resolution and add ColPali as an additional visual search tool.

\paragraph{Inference settings and references.}
Apart from MDocAgent's low reasoning effort, we disable extended reasoning. We use temperature 0 except for DocLens navigator and candidate-answer sampling. We also run gold-page and question-only references with the same GPT-5.4 backbone and answer prompt. We calculate per-question cost from the token usage of every call, rather than only the final answer-generation call.

\section{Evaluation Details}
\label{app:evaluation-protocol}

\subsection{Scoring and Statistical Analysis}
We use Gemini 2.5 Flash as a judge under a fixed rubric. The judge receives the question, reference answer, and model prediction, but not the identity of the answering model or its evidence-access setting. We assign 2 points to fully correct answers, 1 point to partially correct answers, and 0 points to incorrect answers. The rubric accepts paraphrases and equivalent numerical expressions while requiring all requested answer components for full credit. This measure evaluates answer accuracy, not exact string matching or citation faithfulness.

For $N$ evaluated answers with scores $s_i\in\{0,1,2\}$, we report
\begin{equation}
    \mathrm{Accuracy}=\frac{100}{2N}\sum_{i=1}^{N}s_i.
\end{equation}
We estimate 95\% confidence intervals using a document-level cluster bootstrap with 10,000 resamples, preserving the grouping of questions from the same document. We compute paired differences on identical questions.

\subsection{Independent Judge Validation}
Because two evaluated models share the primary judge's model family, we re-judge a stratified sample of 3,323 answers using GPT-5.5 under the same rubric. The sample covers every evaluated model, evidence-access setting, and system. The judges agree exactly on 88.9\% of answers and differ by at most one scoring level on 99.0\%, with quadratic-weighted $\kappa=0.91$. Rankings of the 34 model--setting combinations are also closely aligned (Kendall $\tau=0.91$). In this validation sample, we find no greater leniency from Gemini 2.5 Flash toward Gemini answers.

\section{Computational Resources and Implementation}
\label{app:compute}

\paragraph{Hardware.} All GPU workloads ran on a Slurm cluster with nodes of eight NVIDIA A100 80GB GPUs: open-weight model inference, page retrieval, and LAD-RAG graph construction. Proprietary models, the GPT-5.4-based systems, and the judge were accessed through commercial APIs and required no local GPUs. Across development and test experiments, including pilots and reruns, we used approximately 7,000 A100 GPU-hours, mostly for LAD-RAG graph construction and open-weight model inference.

\paragraph{Software.} We use Python 3.10.12 with PyTorch 2.7 and Hugging Face Transformers 4.53--4.57 for retrieval and graph construction, vLLM for open-weight inference (vLLM 0.9.2 for LAD-RAG graph construction), colpali-engine 0.3.12 for ColPali v1.3 and ColQwen2 v1.0, PyMuPDF 1.26--1.28 for PDF rendering, openpyxl 3.1.5 for table worksheets, and NumPy 2.2.6 and SciPy 1.15.3 for statistics. Pages passed to GPT-5.4 are rendered as 150-DPI PNG images.

\paragraph{Open-weight models and retrieval.} Gemma 3 12B IT and MUSE Glimmer 30B are served with vLLM using a maximum context of 16,384 tokens and one model replica per A100, run data-parallel on up to eight GPUs. Indexing the test documents takes 0.84 A100-hours with ColPali and about 2.4 A100-hours with ColQwen2.

\paragraph{LAD-RAG graph construction.} We build document graphs with Qwen2.5-VL-7B-Instruct served by vLLM at 50\% image scale, using the original description-generation prompts. For the test split, building graphs for 214 documents (21,361 pages) takes 65.8 A100-hours, indexing 1.6 A100-hours, and hosting ColPali for visual search 3.4 A100-hours, for a total of 70.7 A100-hours; the remaining test documents reuse graphs built earlier with the same settings.

\paragraph{API usage.} \Cref{tab:api-usage} reports the mean number of model calls and tokens per question for the GPT-5.4-based systems. SimpleDoc additionally summarizes each page once in advance (1,292 prompt and 452 completion tokens per page on average). The judge scores one answer per request.

\begin{table}[h]
\centering
\small
\setlength{\tabcolsep}{4pt}
\begin{tabular}{@{}lrrr@{}}
\toprule
System (GPT-5.4) & Calls & Prompt tok. & Output tok. \\
\midrule
Question only & 1.0 & 105 & 28 \\
Gold pages & 1.0 & 2,931 & 38 \\
MDocAgent & 5.0 & 14,071 & 639 \\
SimpleDoc & 2.0 & 9,454 & 590 \\
DocAgent & 4.3 & 45,056 & 236 \\
MADQA BM25-MLLM & 4.7 & 9,582 & 137 \\
LAD-RAG & 4.5 & 29,700 & 707 \\
DocLens & 24.7 & 1,321,152 & 2,762 \\
\bottomrule
\end{tabular}
\caption{Mean model calls and tokens per test question for systems with a shared GPT-5.4 backbone.}
\label{tab:api-usage}
\end{table}

\section{Additional Results}
\label{app:additional-results}

\subsection{Full-Document Access Does Not Eliminate the Evidence-Selection Bottleneck}
\label{subsec:allpages}

\begin{figure}[t]
\centering
\includegraphics[width=\columnwidth]{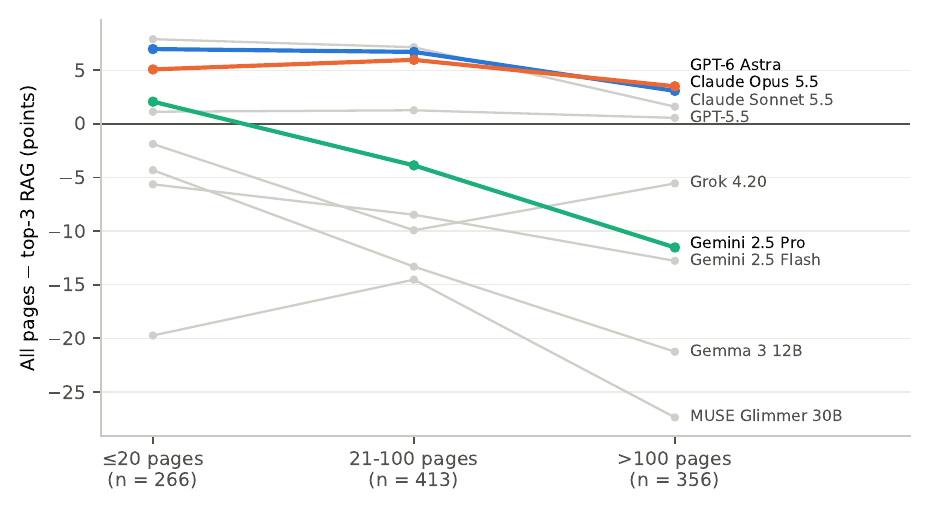}
\caption{All pages minus top-3 RAG accuracy by document length (non-chart questions). Above zero, reading every page beats retrieval; this holds only for the three frontier models (highlighted) and shrinks for long documents.}
\label{fig:allpages}
\end{figure}

Supplying the whole document avoids retrieval errors but does not make models better readers. Reading all pages scores below top-3 retrieval for five of the nine models, by 3.4 (Gemini 2.5 Pro) to 13.3 points (MUSE Glimmer 30B), is indistinguishable for GPT-5.5 (+0.7 [$-$1.1, +2.5]), and helps only GPT-6 Astra, Claude Sonnet 5.5, and Claude Opus 5.5 (+2.8 to +4.3). \Cref{fig:allpages} shows that this advantage depends on document length: the three frontier models gain 5--8 points over RAG on documents of up to 100 pages but only 1.6--3.5 points beyond, while Gemini 2.5 Pro turns negative after 20 pages and reaches $-$11.5 above 100 pages. Even the best all-pages results remain 2.3--4.8 points below gold pages: access to more pages does not substitute for identifying the evidence.

\section{Retrieval misses and contextual distraction}
\label{app:retrieval-distraction}

\begin{figure}[t]
\centering
\includegraphics[width=\columnwidth]{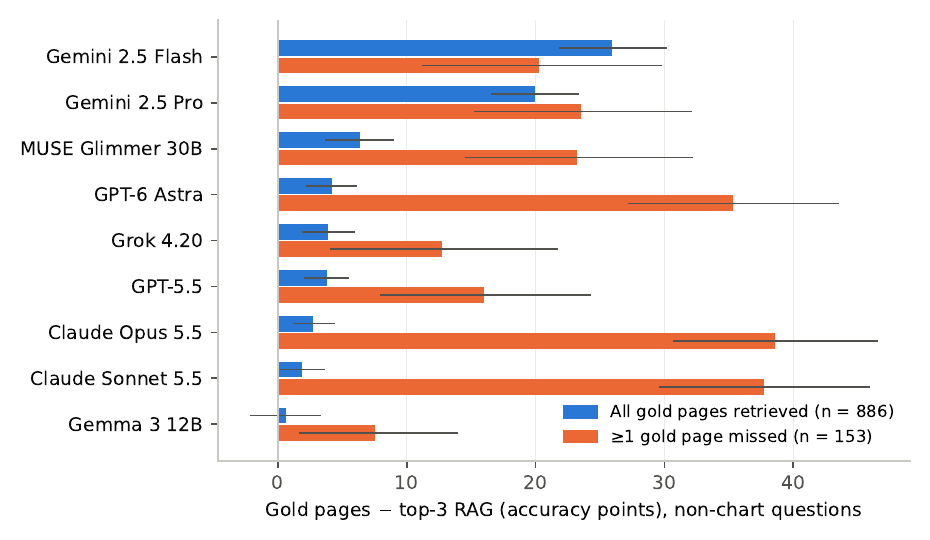}
\caption{Gold$-$RAG accuracy gap on non-chart questions, split by whether all gold pages are in the top three (blue, $n$ = 886) or at least one is missed (orange, $n$ = 153). Strong models lose accuracy only when retrieval misses; Gemini loses accuracy even when every gold page is retrieved. Error bars: 95\% CIs.}
\label{fig:distraction}
\end{figure}

\paragraph{Conditioning on retrieval coverage.}
We analyze the Gold$-$RAG gap according to whether the top three retrieved pages contain all gold pages. \Cref{fig:distraction} shows that the gap narrows to 1.9--4.2 points for the Claude models and GPT-6 Astra when retrieval succeeds, compared with 35--39 points when a gold page is missed. For these models, complete retrieval brings performance close to the gold-page setting, whereas missing evidence produces a much larger deficit.

\paragraph{Errors despite complete evidence.}
The Gemini models exhibit a different pattern. Even when retrieval includes every gold page, Gemini 2.5 Flash loses 25.9 points (95\% CI: $[21.8, 30.2]$) and Gemini 2.5 Pro loses 19.9 points ($[16.5, 23.4]$), corresponding to 83--88\% of their respective overall gaps. In 87\% of these Gemini errors, the model cites the correct gold page but extracts the wrong value. This pattern indicates that retrieving the relevant page is insufficient: additional context can interfere with extracting the answer from that page.

\paragraph{Pipeline controls and gap attribution.}
We use the same runner, prompt, and image resolution in the gold-page and RAG settings. The discrepancy disappears on single-page chart questions, for which both settings receive the same page. These controls support contextual interference, rather than a difference in these pipeline components, as an explanation for the observed deficit. In prompt-matched GPT-5.4 runs, retrieval misses account for 5.1 points of the 8.0-point overall gap (95\% CI: $[5.7, 10.3]$), or 63\%.

\section{Agentic accuracy gains, evidence selection, and cost}
\label{app:agentic-gains-cost}

\subsection{Accuracy gains by system and task}
We restrict the comparison to questions answered by every system and use a shared backbone. Top-3 RAG achieves 74.4\%, full-document input achieves 72.3\%, and gold-page input achieves 78.8\%. Relative to top-3 RAG, SimpleDoc gains 3.5 points (95\% CI: $[1.4, 5.7]$) and MDocAgent gains 2.2 points ($[0.0, 4.2]$). DocLens gains 2.1 points and LAD-RAG gains 1.3 points, but neither improvement is statistically reliable. The gains are concentrated in table questions, with improvements of 6.5--8.4 points for SimpleDoc, MDocAgent, and DocLens; these systems show no corresponding advantage on chart or complex-layout questions. We report LAD-RAG's smaller table-specific gain below. DocAgent underperforms RAG by 4.5 points, while the text-first MADQA agent, which misses values plotted in charts, underperforms by 12.9 points.

\subsection{Adaptive evidence acquisition and graph ablations}
We find that LAD-RAG selects more pages for questions whose supporting evidence spans more pages (Spearman $\rho=0.18$; 95\% CI: $[0.12, 0.23]$). When matched on the number of selected pages, LAD-RAG outperforms ColPali by 3.6 points ($[1.7, 5.4]$) overall and by 14.8 points ($[9.1, 21.1]$) on questions with multi-page evidence. However, its advantage over top-3 RAG is only 1.3 points ($[-0.6, +3.1]$) and is concentrated in table questions ($+4.8$ points). The graph tools and graph context contribute less than one point overall, although graph context provides its largest benefit on complex-layout questions ($+1.9$ points; $[0.4, 3.6]$). These comparisons distinguish the benefit of adaptive evidence selection at a matched page budget from its overall advantage over fixed top-3 retrieval.

\subsection{Cost of additional retrieval and reasoning}
\begin{figure}[t]
\centering
\includegraphics[width=\columnwidth]{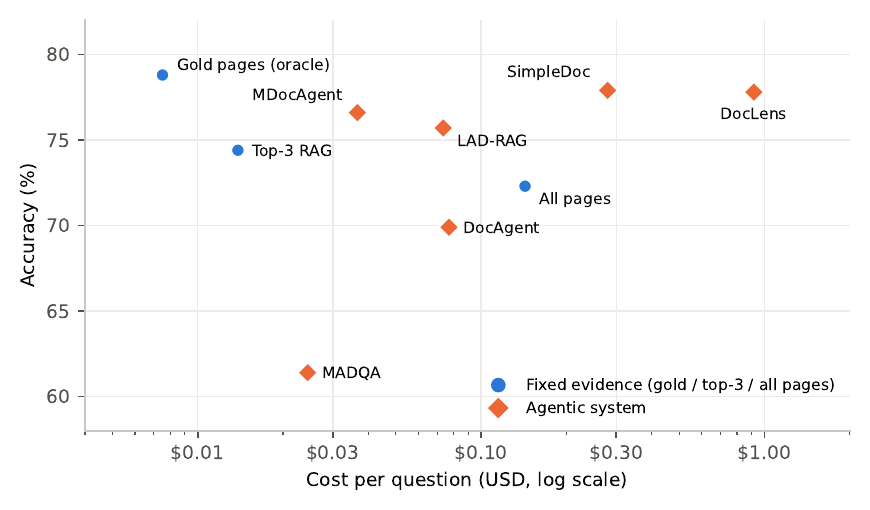}
\caption{Accuracy versus cost per question for GPT-5.4-based systems on the questions answered by every system (log scale; GPT-5.4 list prices). Circles: fixed evidence settings; diamonds: agentic systems.}
\label{fig:cost}
\end{figure}

\Cref{fig:cost} compares per-question costs relative to top-3 RAG. MDocAgent costs $2.6\times$ as much, LAD-RAG $5\times$, SimpleDoc $20\times$, and DocLens $66\times$. SimpleDoc's cost is dominated by its page pre-summaries. Full-document input costs approximately $10\times$ as much as RAG despite its lower aggregate accuracy. In contrast, gold-page input uses a single answer-generation call with perfect evidence and outperforms every system at the lowest cost. Together, these results show that the observed agentic gains come with substantial computational overhead, while accurate evidence selection remains a strong reference for both accuracy and efficiency.

\end{document}